\def\year{2021}\relax
\documentclass[letterpaper]{article} 
\usepackage{aaai21}  
\usepackage{times}  
\usepackage{helvet} 
\usepackage{courier}  
\usepackage[hyphens]{url}  
\usepackage{graphicx} 
\usepackage{natbib}  
\usepackage{caption} 
\usepackage{nicefrac}
\usepackage{amsmath}
\usepackage{amssymb}
\usepackage{multirow}
\usepackage{makecell}

\title{StarNet: towards Weakly Supervised Few-Shot Object Detection}
\author{
    Leonid Karlinsky\thanks{Equal contribution}\hspace{-5pt}*$^{1}$,
Joseph Shtok*$^{1}$,
Amit Alfassy*$^{1,3}$,
Moshe Lichtenstein*$^{1}$,\\
Sivan Harary$^{1}$,
Eli Schwartz$^{1,2}$,
Sivan Doveh$^{1}$,
Prasanna Sattigeri$^{1}$,\\
Rogerio Feris$^{1}$,
Alexander Bronstein$^{3}$,
Raja Giryes$^{2}$ \\
}
\affiliations{
    IBM Research AI$^{1}$, Tel-Aviv University$^{2}$, Technion$^{3}$

}

\begin{document}

\maketitle

\begin{abstract}
Few-shot detection and classification have advanced significantly in recent years. Yet, detection approaches require strong annotation (bounding boxes) both for pre-training and for adaptation to novel classes, and classification approaches rarely provide localization of objects in the scene. In this paper, we introduce StarNet - a few-shot model featuring an end-to-end differentiable non-parametric star-model detection and classification head. Through this head, the backbone is meta-trained using only image-level labels to produce good features for jointly localizing and classifying previously unseen categories of few-shot test tasks using a star-model that geometrically matches between the query and support images (to find corresponding object instances). Being a few-shot detector, StarNet does not require any bounding box annotations, neither during pre-training, nor for novel classes adaptation. It can thus be applied to the previously unexplored and challenging task of Weakly Supervised Few-Shot Object Detection (WS-FSOD), where it attains significant improvements over the baselines. In addition, StarNet shows significant gains on few-shot classification benchmarks that are less cropped around the objects (where object localization is key). 
\end{abstract}
\section{Introduction} \label{sec:intro}
\begin{figure}[ht!]
\vspace{-0.2cm}
\begin{center}
\includegraphics[width=0.475\textwidth]{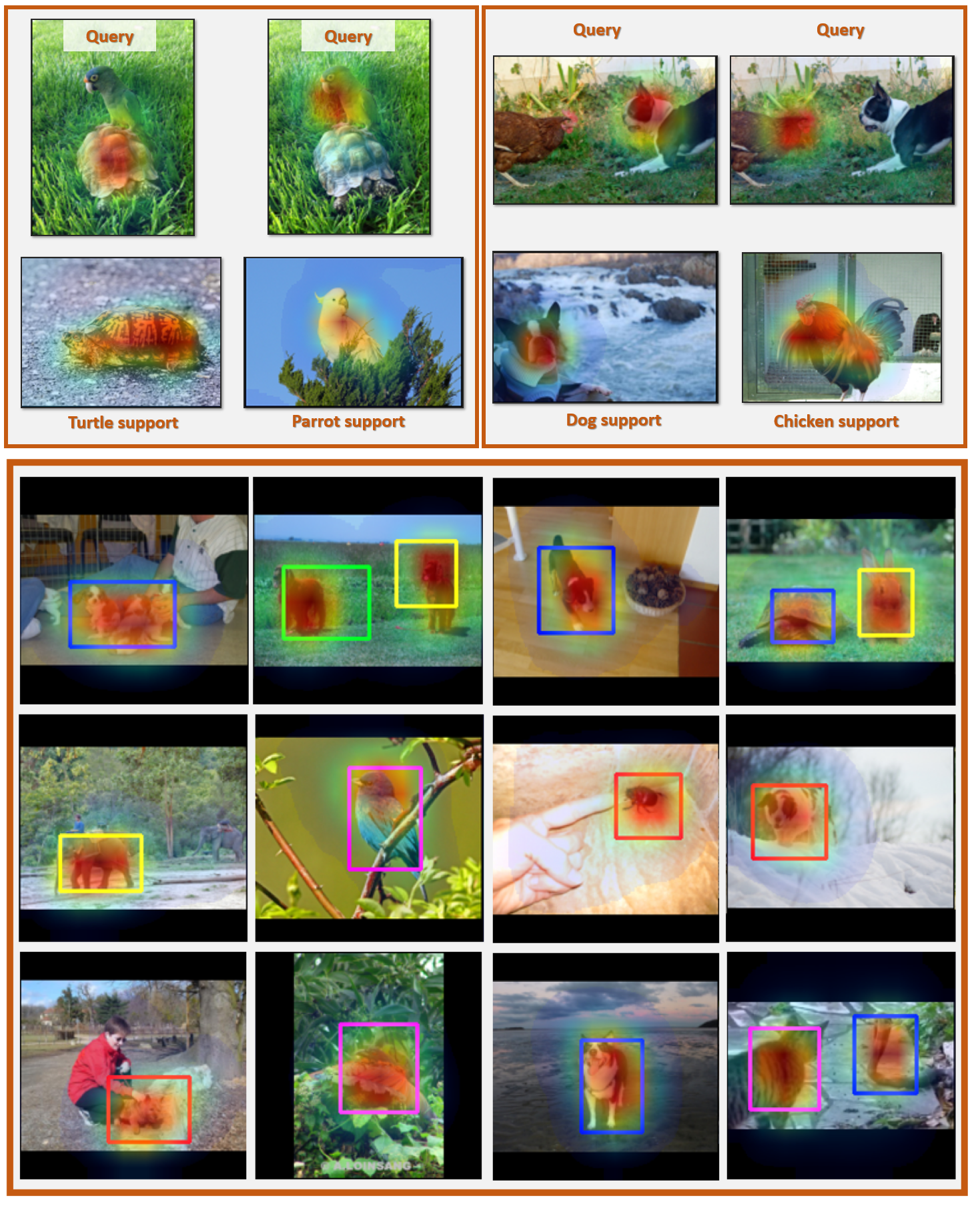}
\end{center}
\caption{StarNet provides evidence for its predictions by finding (semantically) matching regions between query and support images of a few-shot task, thus effectively detecting object instances. \textbf{Top:} Matching regions are drawn as heatmaps for each query and support pair. Clearly, in this situation there is no single correct class label for these queries. Yet, StarNet successfully highlights the matched objects on both the query and the support images, thus effectively explaining the different possible labels. \textbf{Bottom:} StarNet paves the way towards previously unexplored Weakly-Supervised Few-Shot Object Detection (WS-FSOD) task.}
\label{fig:intro}
\vspace{-1.5em}
\end{figure}
\begin{figure*}[!ht]
\begin{center}
   \includegraphics[width=1.0\textwidth]{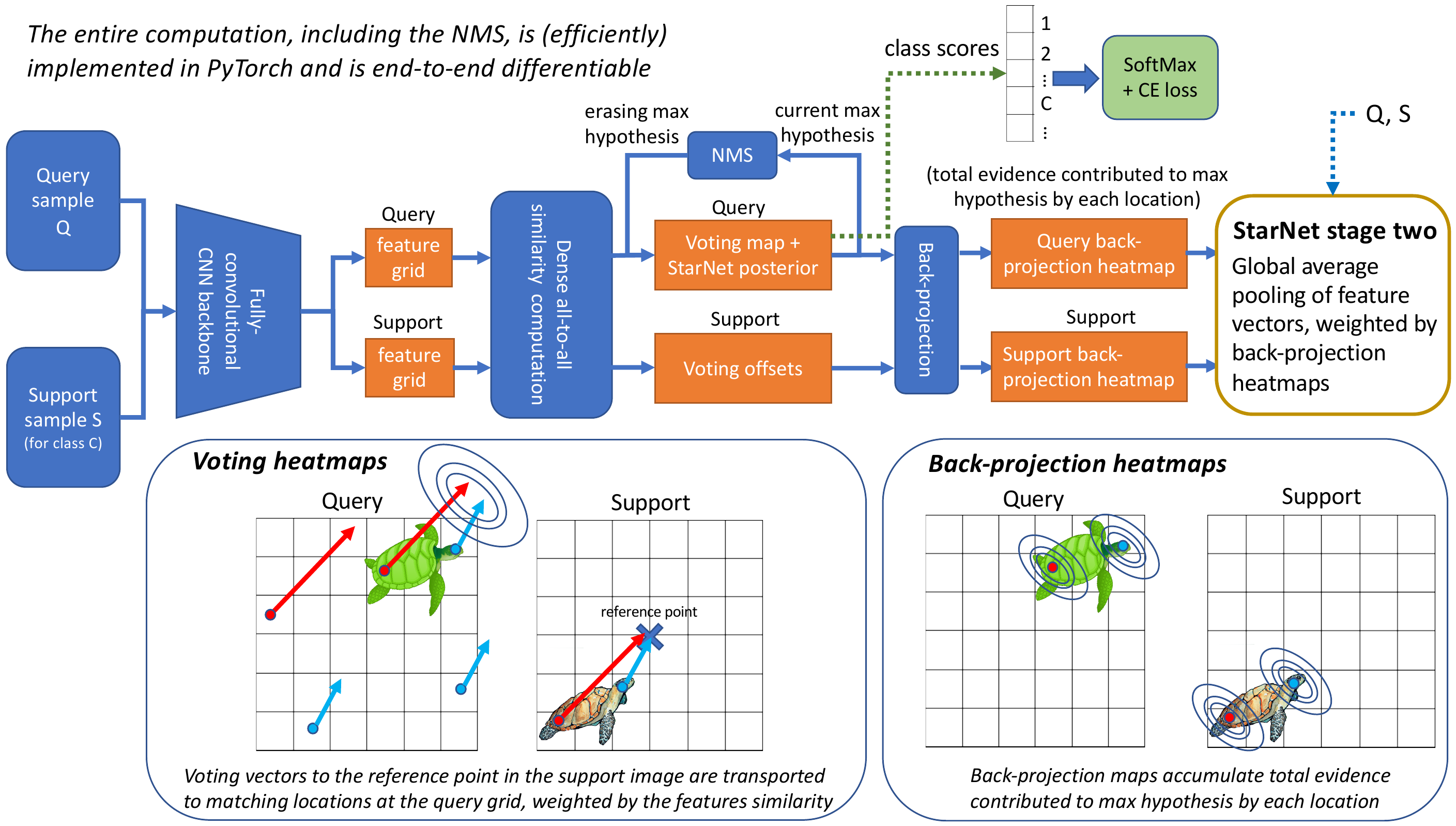}
\end{center}
\caption{\textbf{StarNet overview.} Query image $Q$ is matched to a candidate support image $S$ jointly localizing instances of a shared category (if exist). The NMS iteratively suppresses the max hypothesis allowing to match multiple non-rigid object parts or multiple objects. The back-projection generates decision evidence heatmaps allowing additional refinement stage. StarNet is end-to-end differentiable.}
\label{fig:overview}
\vspace{-1em}
\end{figure*}

Recently, great advances have been made in the field of few-shot learning using deep convolutional neural networks (CNNs). This learning regime targets situations where only a handful of examples for the target classes (typically 1 or 5) are available at test time, while the target classes themselves are novel and unseen during pre-training. Commonly, models are pre-trained on a large labeled dataset of `base' classes, e.g. \cite{Lee2019,Snell2017,Li2017}. There, depending on the application, label complexity varies from image-level class labels (classification), to labeled boxes (detection), to labeled pixel-masks (segmentation). As shown in \cite{closer_look}, few-shot methods are highly sensitive to 'domain shift'. For these methods to be effective, the base classes used for pre-training need to be in the same `visual domain' as the target (test) classes. That said, for applications which require richer annotation, such as detection, entering new visual domains is still prohibitively expensive due to thousands of base classes images that need to be annotated in order to \textit{pre-train} a Few-Shot Object Detector (FSOD), e.g. \cite{lstd2018,Karlinsky2019,Kang2019,Wang2019,Liu2019}, for the new domain. 

\begin{figure*}[!t]
\centering
\includegraphics[width=1.0\textwidth]{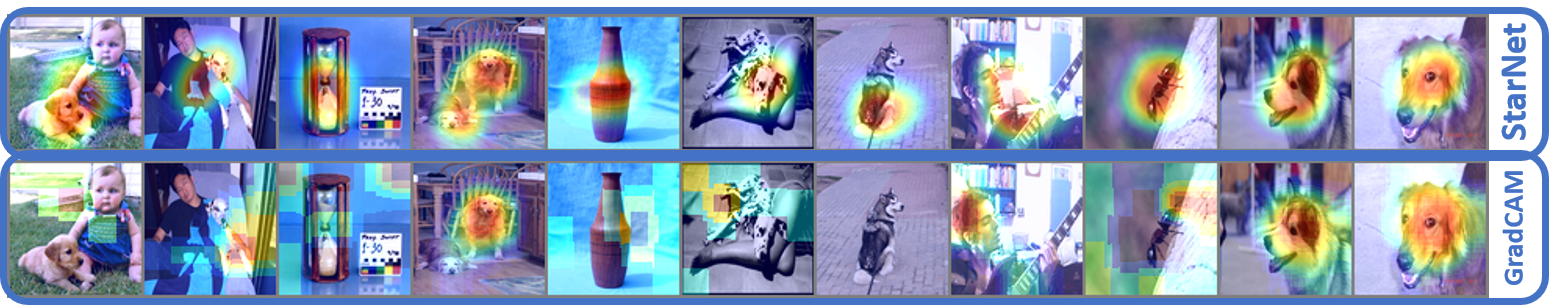}
\caption{
{\bf Comparison with GradCAM:} StarNet back-projection maps (top row) and GradCAM \cite{GradCAM} attention maps (bottom row) computed for MetaOptNet+SVM \cite{Lee2019} on \textit{mini}ImageNet test images. GradCAM failures are likely due to the few-shot setting, or presence of multiple objects.}
\label{fig:grad-cam}
\vspace{-1em}
\end{figure*}

Few-shot classifiers require much less annotation efforts for pre-training, but can only produce image-level class predictions. Of course, general purpose methods such as the popular GradCAM \cite{GradCAM}, are able (to some extent) to highlight the pixels supporting the prediction of any classifier. But, as illustrated in Figure \ref{fig:grad-cam}, and evaluated in Table \ref{table:ws-fs}, these are less effective for few-shot classifiers that need to predict novel classes based on only a few labeled support examples available for a few-shot task. 

In this paper, we introduce a new few-shot learning task: \textit{Weakly-Supervised} Few-Shot Object Detection (WS-FSOD) - pre-training a few-shot detector and adapting it (with few examples) to novel classes \textit{without bounding boxes} and using only image level class label annotations. We also introduce StarNet - a first weakly-supervised few-shot detector that geometrically matches query and support images, classifying queries by localizing objects contained within (Fig. \ref{fig:intro} bottom). StarNet features an end-to-end differentiable head performing non-parametric star-model matching. During training, gradients flowing through the StarNet head teach its underlying CNN backbone to produce features best supporting correct geometric matching. StarNet handles multiple matching hypotheses (e.g. corresponding to multiple objects or object parts), each analyzed by a differentiable back-projection module producing heatmaps of the discovered matching regions (on both query and support images). After training, these heatmaps usually highlight object instances, thus detecting the objects and providing explanations for the model's predictions (Fig. \ref{fig:intro} top).

To summarize, our contributions are as follows: (1) we propose WS-FSOD - a new challenging few-shot learning task of pre-training a few-shot detector and adapting it to novel classes without bounding boxes and using only image class labels; (2) as a solution to WS-FSOD, we propose StarNet - a first end-to-end differentiable non-parametric star-model posed as a neural network, demonstrating promising results for WS-FSOD by significantly outperforming a diverse set of baselines for this new task; (3) as a bonus, not requiring bounding boxes allows StarNet to be directly applied to few-shot classification, where we demonstrate it to be especially useful on benchmarks in which images are less cropped around the objects (e.g. CUB and ImageNetLOC-FS), and for which object localization is key.
\section{Related Work} \label{sec:related}
In this section we briefly review the modern few-shot learning focusing on meta-learning, discuss weakly-supervised detection, cover star-model related methods, and review methods for few-shot localization and detection.

\textbf{Meta-learning} methods \cite{Vinyals2016,Snell2017,relationnet,Li2019,Finn2017,Li2017,Zhou2018,Ravi2017,Munkhdalai2017,leo,closer_look,Schwartz2018,Oreshkin2018,Zhang2019,Zhang2019a,Gidaris2019a} learn from few-shot tasks (or episodes) rather then from individual labeled samples. Such tasks are small datasets, with few labeled training (support) examples, and a few test (query) examples. The goal is to learn a model that can adapt to new tasks with novel categories, unseen during training.  In \cite{Dvornik2019} ensemble methods for few-shot learning are evaluated. MetaOptNet \cite{Lee2019} utilizes an end-to-end differentiable SVM solver on top of a CNN backbone. \cite{Gidaris2019} combines few-shot supervision with self-supervision, in order to boost the few-shot performance. In \cite{Qiao2019,Li2019_LST,Kim2019,Gidaris2019} additional unlabeled data is used, while \cite{am3} leverages additional semantic information available for the classes.

\textbf{Star Models (SM)} and \textbf{Generalized Hough Transform (GHT)} techniques were popular classification and detection methods before the advent of CNNs. In these techniques, objects were modeled as a collection of parts, independently linked to the object variables via Gaussian priors to allow local deformations. Classically, parts were represented using patch descriptors \cite{Sali1999,Leibe2004,maji2009,Karlinsky2017}, or SVM part detectors in DPM \cite{Felzenszwalb2009}. DPM was later extended to CNN based DPM in \cite{Girshick2014}. Recently, in \cite{Qi2019} GHT was used to detect objects in 3D point clouds, in the fully supervised and non-few-shot setting. Unlike DPM \cite{Felzenszwalb2009,Girshick2014}, StarNet is non-parametric, in a sense that parts are not explicitly learned and are not fixed during inference, and unlike all of the aforementioned methods \cite{Sali1999,Leibe2004,maji2009,Felzenszwalb2009,Girshick2014,Karlinsky2017,Qi2019}, it is trained using only class labels (no bounding boxes) and targets the few-shot setting. In \cite{Lin2017a} a \textit{non} few-shot classification network is trained through pairwise local feature matching, but unlike in StarNet, no geometrical constraints on the matches are used. Finally, unlike the classical approaches \cite{Sali1999,Leibe2004,maji2009,Felzenszwalb2009}, StarNet features (used for local matching) are not handcrafted, but are rather end-to-end optimized by propagating gradients through StarNet head to a CNN backbone.

\textbf{Weakly-supervised object detection} refers to techniques that learn to detect objects despite being trained with only image-level class labels. In a number of works, an external region proposal mechanism (e.g., Selective Search \cite{selective_search}) is employed to endow a pre-trained CNN classifier with a detection head \cite{Bilen2016}, or to provide initial proposals for RPN \cite{Zeng2019}. In \cite{PCL}, the proposals are clustered into groups to facilitate iterative training of instance classifiers. More recently, in \cite{Tang2019}, a region proposal sub-network is trained jointly with the backbone, by refining initial (sliding window) proposals. To the best of our knowledge, no prior works have considered the weakly-supervised detection in the \textit{few-shot} setting. 

\textbf{Few-shot with localization and attention} is a relatively recent research direction. Unlike StarNet, most of these methods rely on bounding box supervision during pre-training. Using bounding boxes, several works \cite{lstd2018,Karlinsky2019,Kang2019,Wang2019,Liu2019,Wang2020} have extended object detection techniques \cite{Ren2015,Liu2016b} to few-shot setting. \cite{Wertheimer2019} uses an attention module trained using bounding boxes. SILCO \cite{Hu2019} trains using bounding boxes to localizes objects in $1$-way / $5$-shot mode only. \cite{Shaban2019} uses Multiple Instance Learning and an RPN pre-trained using bounding boxes on MS-COCO \cite{Lin2014}. SAML \cite{Hao2019} and DeepEMD \cite{Zhang2020a} compute a dense feature matching applying MLP or EMD metric as a classifier, but unlike StarNet geometric matching is not employed. In CAN \cite{hou2019a} attention maps for query and support images are generated by $1\times1$ convolution applied to a pairwise local feature comparison map. These attention maps are not intended for object localization, so unlike StarNet, geometry of the matches in \cite{hou2019a} is not modeled. In DC \cite{lifchitz2019a} a classifier is applied densely on each of the local features in the feature map, their decisions are globally averaged, unlike StarNet, without employing geometry.

Recently, \cite{Choe2020} proposed a few-shot protocol for Weakly Supervised Object \textit{Localization} (WSOL) - given (a single) true class label of the test image, localizing an object of that class in it. In their protocol \textit{ImageNet pre-trained} models are fine-tuned using $5$-shots with \textit{bounding boxes} supervision. In contrast, in this paper we propose Weakly Supervised Few-Shot Object \textit{Detection} (WS-FSOD) protocol, where: test images (potentially multiple) class labels are not given; models are pre-trained from scratch on the train portions of the benchmarks and adapted to novel classes using $1$ or $5$ shots; and no bounding boxes are used for training. We believe our WS-FSOD protocol to be more fitting situations of entering a new visual domain, where ImageNet-scale pre-training and box annotations are not available.
\section{Method} \label{sec:method}
Here we provide the details of the StarNet method. First we describe the approach for calculating the StarNet posterior for each query-support pair and using it to predict the class scores for every query image in a single-stage StarNet. Next we explain how to revert StarNet posterior computation using back-projection, obtaining evidence maps (on both query and support) for any hypothesis. Then we show how to enhance StarNet performance by adding a second-stage hypothesis classifier utilizing the evidence maps to pool features from the (query and support) matched regions, effectively suppressing background clutter. Finally, we provide implementation details and running times. Figure \ref{fig:overview} gives an overview of our approach. 

\subsection{Single-stage StarNet} \label{sec:single-stage-starnet}
StarNet is trained in a meta-learning fashion, where $k$-shot, $n$-way training episodes are randomly sampled from the train data. Each episode (a.k.a few-shot task) $E$ consists of $k$ random support samples ($k$-shot) and $q$ random query samples for each of $n$ random classes ($n$-way). Denote by $Q$ and $S$ a pair of query and support images belonging to $E$. Let $\phi$ be a fully convolutional CNN feature extractor, taking a square RGB image input and producing a feature grid tensor of dimensions $r\times r\times f$ (here $r$ is the spatial dimension, and  $f$ is the number of channels). Applying $\phi$ on $Q$ and $S$ computes the query and support grids of feature vectors: 
\begin{equation}
\begin{split}
&\{ \phi(Q)_{i,j} \in \mathcal{R}^{f} |\; 1 \le i, j \le r \} \\
&\{ \phi(S)_{l,m} \in \mathcal{R}^{f} |\; 1 \le l, m \le r \}
\end{split}
\end{equation} 
For brevity we will drop $\phi$ in further notation and write $Q_{i,j}$ and $S_{l,m}$ instead of $\phi(Q)_{i,j}$ and $\phi(S)_{l,m}$. We first $L_2$ normalize $Q_{i,j}$ and $S_{l,m}$ for all grid cells, and then compute a tensor $D$ of size $r \times r \times r \times r$ of all pairwise distances between $Q$ and $S$ feature grids cells: 
\begin{equation}
    D_{i,j,l,m} = ||Q_{i,j}-S_{l,m}||^2
\end{equation}
$D$ is efficiently computed for all support-query pairs simultaneously via matrix multiplication with broadcasting. We then convert $D$ into a (same size) tensor of unnormalized probabilities $P$, where: 
\begin{equation}
    P_{i,j,l,m}=e^{-0.5 \cdot D_{i,j,l,m}/ \sigma_{f}^{2}}
\end{equation}
is the probability that $Q_{i,j}$ matches $S_{l,m}$ in a sense of representing the same part of the same category. Some object part appearances are more rare than others; to accommodate for that, $P$ is normalized to obtain the tensor $R$ of the same size, where 
$R_{i,j,l,m} = P_{i,j,l,m} / N_{i,j}$ 
is the likelihood ratio between `foreground' match probability $P_{i,j,l,m}$, and  the `background' probability $N_{i,j}$ of 'observing' $Q_{i,j}$ in a random image, approximated as: 
\begin{equation}
N_{i,j}=\sum_{S}\sum_{l,m}P_{i,j,l,m}
\end{equation}
where $\sum_{S}$ is computed by matching the same query $Q$ to all of the supports in the episode. Note that in $R_{i,j,l,m}$, the normalization factor of unnormalized probabilities $P$ cancels out. Let $w=(r/2,r/2)$ be a reference point in the center of $S$ feature grid. We compute voting offsets as  $o_{l,m} = w - (l,m)$ and the voting target as $t_{i,j,l,m} = (i,j) + o_{l,m}$ being the corresponding location to the reference point $w$ on the query image $Q$ assuming that indeed $Q_{i,j}$ matches $S_{l,m}$. By construction, $t_{i,j,l,m}$ can be negative, with values ranging between $(-r/2,-r/2)$ and $(3r/2,3r/2)$, thus forming a $2r \times 2r$ hypothesis grid of points in coordinates of $Q$ potentially corresponding to point $w$ on $S$.

Next, for every point $(x,y)$ on the hypothesis grid of $Q$, StarNet accumulates the overall belief $A(x,y)$ that $(x,y)$ corresponds to $w$ (on $S$) considering independently the evidence $R_{i,j,l,m}$ from all potential matches between support and query features. In probabilistic sense, $A(x,y)$ relates to Naive-Bayes \cite{Bishop2006PatternLearning}, and hence should accumulate \textit{log}-likelihood ratios $log(R_{i,j,l,m})$. However, as in \cite{Karlinsky2017}, to be more robust to background clutter, in StarNet, likelihood ratios are directly accumulated:
\begin{equation}
A(x,y) = \sum_{\substack{\{i,j,l,m\}\; s.t.\\ t_{i,j,l,m}=(x,y)}} R_{i,j,l,m}
\end{equation}
For each hypothesis $(x,y)$, the final StarNet posterior $V_{Q,S}(x,y)$ is computed by convolving $A$ with $G(\sigma_g)$ - a symmetric Gaussian kernel:
    $V_{Q,S} = G(\sigma_g) \circledast A$.
This efficiently accounts for any random relative location shift (local object part deformation) allowed to occur with the $G(\sigma_g)$ Gaussian prior for any matched pair of $Q_{i,j}$ and $S_{l,m}$.

We compute the score (logit) of predicting the category label $c$ for $Q$ as:
\begin{equation}
    SC_1(c ; Q) = \frac{1}{k} \cdot \sum_{\substack{S \in E\;s.t.\\C(S)=c}} \max_{x,y} V_{Q,S}(x,y)
\end{equation}
where $C(S)$ is the class label of $S$, and $k$ is the number of shots (support samples per class) in the episode $E$. During meta-training the CNN backbone $\phi$ is end-to-end trained using Cross Entropy (CE) loss between $SC_1(c ; Q)$ (after softmax) and the known category label of $Q$ in the training episode. The need to only match images with the same class label, drives the optimization to maximally match the regions that correspond to the only thing that is in fact shared between such images - the instances of the shared category (please see Appendix for examples and video illustrations).

\subsection{Back-projection maps} \label{sec:backprojection}
For any pair of query $Q$ and support $S$, and any hypothesis location $(\hat{x},\hat{y})$ on the $2r \times 2r$ grid, and in particular one with the maximal StarNet posterior value $(\hat{x},\hat{y}) = \arg\max_{x,y} V_{Q,S}(x,y)$, we can compute two back-projection heatmaps (one for $Q$ and one for $S$). These are $r \times r$ matrices in the feature grid coordinates of $Q$ and $S$ respectively, whose entries contain the amount of contribution that the corresponding feature grid cell on $Q$ or $S$ gave to the posterior probability $V_{Q,S}(\hat{x},\hat{y})$:
\begin{equation}
    BP_{Q|S}(i,j) = \sum_{l,m} R_{i,j,l,m} \cdot e^{-0.5 \cdot ||t_{i,j,l,m}-(\hat{x},\hat{y})||^2 / \sigma_{g}^{2}} \;\;\;\;
\end{equation}
the $BP_{S|Q}(l,m)$ is computed in completely symmetrical fashion by replacing summation by $l,m$ with summation by $i,j$. After training, the back-projection heatmaps are highlighting the matching regions on $Q$ and $S$ that correspond to the hypothesis $(\hat{x},\hat{y})$, which for query-support pairs sharing the same category label are in most cases the instances of that category (examples provided in Appendix).

The back-projection is iteratively repeated by suppressing $(\hat{x},\hat{y})$ (and its $3 \times 3$ neighborhood) in $V_{Q,S}(x,y)$ as part of the Non-Maximal Suppression (NMS) process implemented as part of the neural network. NMS allows for better coverage of non-rigid objects detected as sum of parts and for discovering additional objects of the same category. Please see Fig. \ref{fig:intro}, Fig. \ref{fig:grad-cam} (image $4$, top row), and the Appendix, for examples of images with multiple objects detected by StarNet. In our implementation, NMS repeats until the next maximal point is less then an $\eta=0.5$ from the global maximum.

\subsection{Two-stage StarNet} \label{sec:two-stage-starnet}
Having computed the $BP_{Q|S}$ and $BP_{S|Q}$ back-projection heatmaps, we take inspiration from the 2-stage CNN detectors (e.g. FasterRCNN \cite{Ren2015}) to enhance the StarNet performance with a second stage classifier benefiting from category instances localization produced by StarNet (in $BP_{Q|S}$ and $BP_{S|Q}$). We first normalize each of the $BP_{Q|S}$ and $BP_{S|Q}$ to sum to 1, and then generate the following pooled feature vectors by weighted global average pooling with $BP_{Q|S}$ and $BP_{S|Q}$ weights:
\begin{equation}
\begin{split}
    F_{Q|S} &= \sum_{i,j} BP_{Q|S}(i,j) \cdot Q_{i,j} \\
    F_{S|Q} &= \sum_{l,m} BP_{S|Q}(l,m) \cdot S_{l,m}
\end{split}
\end{equation}
here the feature grids $Q_{i,j}$ and $S_{l,m}$ can be computed using $\phi$ or using a separate CNN backbone trained jointly with the first stage network (we evaluate both in experiments section). Our second stage is a variant of the Prototypical Network (PN) classifier \cite{Snell2017}. We compute the prototypes for class $c$ and the query $Q$ as: 
\begin{equation}
\begin{split}
    F^P_{c|Q} &= \frac{1}{k} \cdot \sum_{S \in E\;s.t.\;C(S)=c} F_{S|Q} \\
    F^P_{Q|c} &= \frac{1}{k} \cdot \sum_{S \in E\;s.t.\;C(S)=c} F_{Q|S}
\end{split}
\end{equation}
Note that as opposed to PN, our query ($F^P_{Q|c}$) and class ($F^P_{c|Q}$) prototypes are different for each query + class pair. Finally, the score of the second stage classifier for assigning label $c$ to the query $Q$ is:
\begin{equation}
    SC_2(c;Q) = -||F^P_{Q|c}-F^P_{c|Q}||_2
\end{equation}
The predictions of the classifiers of the two stages of StarNet are fused using geometric mean to compute the joint prediction as ($sm=softmax$):
\begin{equation}
    SC(c;Q) = \sqrt{sm(SC_1(c;Q)) \cdot sm(SC_2(c;Q))}
\end{equation}

\subsection{Implementation details} \label{sec:implementation-details}
Our implementation is in PyTorch $1.1.0$ \cite{Paszke2017}, and is based on the public code of \cite{Lee2019}. In all experiments the CNN backbone is ResNet-$12$ with 4 convolutional blocks (in $2$-stage StarNet we evaluated both single shared ResNet-$12$ backbone and a separate ResNet-$12$ backbone per stage). To increase the output resolution of the backbone we reduce the strides of some of its blocks. Thus, for benchmarks with $84 \times 84$ input image resolution, the block strides were $[2,2,2,1]$ resulting in $10 \times 10$ feature grids, and for $32 \times 32$ input resolution (in Appendix), we used $[2,2,1,1]$ strides resulting in $8 \times 8$ feature grids. This establishes naturally the value for $r$, we intend to explore other values in future work.
We use four $1$-shot, $5$-way episodes per training batch, each episodes with $20$ queries. The hyper-parameters $\sigma_{f}=0.2$, $\sigma_{g}=2$, and $\eta=0.5$ were determined using validation. As in \cite{Lee2019}, we use $1000$ batches per training epoch, $2000$ episodes for validation, and $1000$ episodes for testing. We train for $60$ epochs, changing our base $LR=1$ to $0.06, 0.012, 0.0024$ at epochs $20, 40, 50$ respectively. The best model for testing is determined by validation.
On a single NVidia K40 GPU, our running times are: $1.15$s/batch in 1-stage StarNet training; $2.2$ s/batch in 2-stage StarNet training (in same settings \cite{Lee2019} trains in $2.1$s/batch); and $0.01$s per query in inference. GPU peak memory was $\sim 30$MB per image.
\section{Experiments} \label{sec:experiments}
In all of experiments, only the class labels were used for training, validation, and for the support images of the test few-shot tasks. The bounding boxes were used only for performance evaluation.
For each dataset we used the standard train / validation / test splits, which are completely disjoint in terms of contained classes. Only episodes generated from the training split were used for meta-training; the hyper-parameters and the best model were chosen using the validation split; and test split was used for measuring performance. Results on additional datasets are provided in Appendix.
\begin{table*}[!ht]
	\centering
	\caption{\textbf{WS-FSOD performance:} comparing to baselines, performance measured in Average Precision (AP\%). GC = GradCAM, SS = SelectiveSearch. RepMet \cite{Karlinsky2019} and TFA \cite{Wang2020} are fully-supervised upper bounds. $^{(1)}$using official code and best hyper-parameters between defaults and those found by tuning on val. set for each benchmark.}
	\begin{tabular}{l|l|cc|cc}
		\multicolumn{2}{c}{}&
     \multicolumn{2}{c}{1-shot}  & \multicolumn{2}{c}{5-shot}  \\
    dataset & method   & $IoU\geq0.3$ & $IoU\geq0.5$  &   $IoU\geq0.3$ & $IoU\geq0.5$ \\
    \hline
      \shortstack{\textbf{Imagenet LOC-FS}} & RepMet (fully supervised \textit{upper bound}) &
      \quad 59.5$^{(1)}$      & 56.9      &    \quad 70.7$^{(1)}$  &68.8    \\
     &
    \shortstack{MetaOpt$^{(1)}$+GC} & 32.4    & 13.8        &  51.9   &   22.1     \\
     & \shortstack{MetaOpt$^{(1)}$+SS} & 16.1      & 4.9       &  27.4   &  10.2    \\

    & \shortstack{PCL$^{(1)}$ \cite{PCL}} & 25.4      & 9.2      &  37.5  &   11.3    \\
    & \shortstack{CAN$^{(1)}$ \cite{hou2019a}} & 23.2          & 10.3      &  38.2   &   12.7    \\
    & \shortstack{random+StarHead} & 2.1          & 0.6      &  3.6   &   0.8    \\
    & \shortstack{pretrained+StarHead} & 22.9          & 10.2      &  31.0   &   21.3    \\

    & \textbf{StarNet (ours)}  & \textbf{50.0}       & \textbf{26.4}      &  \textbf{63.6}    &   \textbf{34.9}    \\
    \hline
    \textbf{CUB} &
        \shortstack{MetaOpt$^{(1)}$+GC} & 53.3          & 12.0        &  72.8   &   14.4    \\
        & \shortstack{MetaOpt$^{(1)}$+SS} & 19.4      & 6.0        &  26.2   &   6.4    \\

       & \shortstack{PCL$^{(1)}$ \cite{PCL}} & 29.1      & 11.4       &  41.1  &   14.7   \\
        & \shortstack{CAN$^{(1)}$ \cite{hou2019a}} & 60.7          & 19.3        &  74.8   &   26.0   \\
        & \shortstack{random+StarHead} & 3.5          & 0.6      &  6.0   &   0.9    \\
        & \shortstack{pretrained+StarHead} & 47.6          & 13.2      &  62.2   &   17.3    \\
        & \textbf{StarNet (ours)}  & \textbf{77.1}        & \textbf{27.2}        &  \textbf{86.1}   &   \textbf{32.7}    \\
    \hline
    \textbf{Pascal VOC} & 
        TFA (fully-supervised \textit{upper bound}) & - & 31.4 & - & 46.8\\
        (average over $5$-way sets) & StarNet (ours) & 34.1 & 16.0 & 52.9 & 23.0 \\
    
		\hline
	\end{tabular}
    \label{table:ws-fs}
    \vspace{-1em}
\end{table*}

The \textbf{CUB} fine-grained dataset \cite{Welinder2010} consists of $11,788$ images of birds of $200$ species. We use the standard train, validation, and test splits, created by randomly splitting the $200$ species into $100$ for training, $50$ for validation, and $50$ for testing and used in all few-shot works. All images are downsampled to $84 \times 84$. Images are \textit{not cropped} around the birds, which appear on cluttered backgrounds.

The \textbf{ImageNetLOC-FS} dataset \cite{Karlinsky2019} contains $331$ animal categories from ImageNetLOC \cite{imagenet} split into: $101$ for train, $214$ for test, and $16$ for validation. Since animals are typically photographed from afar, and as the images in this dataset are pre-processed to $84 \times 84$ square size with aspect ratio preserving padding (thus adding random padding boundaries), commonly images in this dataset are \textit{not cropped} around the objects (some examples are in figure \ref{fig:intro} bottom).

%
%
\begin{table*}[!t]
  \centering
    \caption{Few-shot classification accuracy ($\%$), for all methods the $0.95$ confidence intervals are $<1\%$ (omitted for brevity). For fair comparison, showing only results that \textit{do not} use the validation set for training, \textit{do not} use the transductive or semi-supervised setting, use standard input resolution $84 \times 84$, 
    and \textit{do not use} additional information such as class label or class attributes embedding. Results on additional few-shot classification benchmarks are provided in Appendix.
    $^{(1)}$Results from \cite{closer_look}, best result among resnet-10/18/34. 
    $^{(2)}$using official code and best hyper-parameters between defaults and those found by tuning on validation set for each benchmark.
    $^{(3)}$ shared backbone between StarNet stage-1 and stage-2.
  }
  \begin{tabular}{l|l|cc|cc}
      \multicolumn{2}{l}{} &
      \multicolumn{2}{c}{\textbf{ImageNetLOC-FS}} &
      \multicolumn{2}{c}{\textbf{CUB}} \\ 
      \hline
  \textbf{method} & 
      \textbf{backbone architecture} & 
      \textbf{1-shot} & \textbf{5-shot} & 
      \textbf{1-shot} & \textbf{5-shot} 
      \\
   \hline
   SAML \cite{Hao2019} & conv4
                  & - & - 
                  & 69.35 & 81.56
                \\
   Baseline$^{(1)}$ \cite{closer_look}  & resnet-34 & - & - & 67.96 & 84.27 
   \\
    Baseline++$^{(1)}$ \cite{closer_look} & resnet-34 & - & - & 69.55 & 85.17 
    \\
    MatchingNet$^{(1)}$ \cite{Vinyals2016}   & resnet-34 & - & - & 73.49 & 86.51 
    \\
    ProtoNet$^{(1)}$ \cite{Snell2017}   & resnet-34 & - & - & 73.22 & 87.86 
    \\
    MAML$^{(1)}$ \cite{Finn2017}   & resnet-34 & - & - & 70.32 & 83.47 
    \\
    RelationNet$^{(1)}$ \cite{relationnet}   & resnet-34 & - & - & 70.47 & 84.05 
    \\
   Dist. ensemble \cite{Dvornik2019} &
                  ensemble of $20\times$ resnet18
                  & - & - 
                  & 70.07 & 85.2 
                  \\       
   $\Delta$-encoder \cite{Schwartz2018} &
                resnet-18
                & - & - 
                & 69.80 & 82.60
                \\       
   DeepEMD \cite{Zhang2020a}
                  & resnet-12
                  & - & -
                  & 75.65 & 88.69 \\
   CAN \cite{hou2019a}
                & resnet-12
                & 57.1\tiny{$^{(2)}$} & 73.9\tiny{$^{(2)}$} 
                & 75.01\tiny{$^{(2)}$} & 86.8\tiny{$^{(2)}$}
                \\       
   MetaOpt \cite{Lee2019}
                  & resnet-12
                  & 57.7\tiny{$^{(2)}$} & 74.8\tiny{$^{(2)}$} 
                  & 72.75\tiny{$^{(2)}$} & 85.83\tiny{$^{(2)}$}
                  \\
   \hline
   \textbf{StarNet - shared backbone (ours)}\tiny{$^{(3)}$}
                  & resnet-12
                  & 61.0 & 77.0 
                  & 79.44 &  88.8 
                  \\        
   \textbf{StarNet (ours)}
                  & $2\times$ resnet-12 = resnet-18
                  & \textbf{63.0} & \textbf{78.0} 
                  & \textbf{79.58} &  \textbf{89.5} 
                  \\        
\end{tabular}
  \label{table:results_classification}
   \vspace{-1em}
\end{table*}
\subsection{Weakly-Supervised Few-Shot Object Detection} \label{sec:weakly-supervised-few-shot-detection}
We used ImageNetLOC-FS and CUB few-shot datasets, as well as PASCAL VOC \cite{Everingham2010} experiment from \cite{Wang2020}, to evaluate StarNet performance on the proposed WS-FSOD task. All datasets have bounding box annotations, that in our case were used \textit{only} for evaluating the detection quality. The ImageNetLOC-FS and the PASCAL VOC experiments allow comparing StarNet's performance directly to \textit{Fully-Supervised} FSOD SOTA: RepMet \cite{Karlinsky2019} and TFA \cite{Wang2020} respectively, both serving as a natural performance upper bound for the \textit{Weakly-Supervised} StarNet. 
Since, to the best of our knowledge, StarNet is the first method proposed for WS-FSOD, we also compare its performance to a wide range of weakly-supervised baselines. 

Two baselines are based on a popular few-shot classifier MetaOpt \cite{Lee2019} combined with GradCAM or SelectiveSearch \cite{selective_search} for localizing the classified categories. Third baseline is PCL \cite{PCL} - recent (non few-shot) WSOD method. Using official PCL code, we pre-trained it on the same training split as used for training StarNet, and adapted it by finetuning on support set of each of the test few-shot tasks. Fourth is the SOTA attention based few-shot method of CAN \cite{hou2019a}, that also has some ability to localize the objects.
Finally, as a form of ablation, we offer two baselines evaluating the (non-parametric) StarNet head on top of ResNet-12 backbone that is: (i) randomly initialized, or (ii) pre-trained using a linear classifier. These baselines underline the importance of training the backbone end-to-end \textit{through StarNet head} for the WS-FSOD higher gains. 
The results for WS-FSOD experiments and comparisons (averaged over $500$ $5$-way test episodes) are summarized in Table \ref{table:ws-fs}, and qualitative examples of StarNet detections are shown in Figure \ref{fig:intro}(bottom). 

For all methods and FS-WSOD experiments, we use the standard detection metric where detected bounding box is considered correct if its Intersection-over-Union (IoU) with a ground truth box is above threshold and its top-scoring class prediction is correct. We report Average Precision (AP) under this metric using $0.3$ and $0.5$ IoU thresholds. For all methods producing heatmaps, the bounding boxes were obtained using the CAM algorithm from \cite{CAM_original,Zhang2002} (as in most WSOD works).

StarNet results are higher by a large margin than results obtained by all the compared baselines. This is likely due to StarNet being directly end-to-end optimized for classifying images by detecting the objects within (using the proposed star-model geometric matching), while the other methods are either: not intended for few-shot (PCL), or optimized attention for classification and not for detection (CAN), or intended for classification and not detection (MetaOpt) - which cannot be easily bridged using the standard techniques for localization in classifiers (GradCAM, SelectiveSearch).
As can be seen from Table \ref{table:ws-fs}, for $IoU \geq 0.3$ the StarNet is close to the fully supervised few-shot RepMet detector with about $10$ AP points gap in $1$-shot and about $7$ points gap in $5$-shot. However, the gap increases substantially for $IoU \geq 0.5$. We suggest that this gap is mainly due to partial detections (bounding box covering only part of an object) - a common issue with most WSOD methods. 
Analysis corroborating this claim is provided in the Appendix.

Finally, we performed \cite{Wang2020}'s few-shot PASCAL VOC evaluation (three $5$-way novel category sets), comparing to the fully-supervised (with boxes) SOTA FSOD method TFA proposed in that paper (Table \ref{table:ws-fs} bottom). As TFA uses a (ResNet-101) backbone pre-trained on ImageNet (as common in FSOD works), in this experiment we used StarNet pre-trained on ImageNetLOC-FS (weakly supervised, without boxes) excluding PASCAL overlapping classes. Consistently with comparison to RepMet upper bound, under a more relaxed boxes tightness requirement of $IoU \geq 0.3$ (as discussed, used mostly due to partial detections), the AP of the weakly-supervised StarNet is close to the fully supervised TFA upper bound. Qualitative results from PASCAL experiment are provided in the Appendix.

\subsubsection{Limitations}
StarNet detects multiple objects of different classes on the same query image via matching to different support images. It can also detect multiple instances of the same class via its (differentiable) NMS if their back-projection heatmap blobs are non-overlapping or if they are matched to different support images for that class. Yet in some situations, if same class instances are overlapping on the query image and are matched to the same support image (as is bound to happen in $1$-shot tests) - they would be detected as a single box by StarNet. Enhancing StarNet to detect overlapping instances of the same class is beyond the scope of this paper and an interesting future work direction.

\subsection{Few-shot classification} \label{sec:few-shot-classification}
StarNet is a WS-FSOD, trainable just from image class labels, and hence is readily applicable to standard few-shot classification testing. We used the standard few-shot classification evaluation protocol, exactly as in \cite{Lee2019}, using $1000$ random $5$-way episodes, with $1$ or $5$ shots. 
StarNet is optimized to classify the images \textit{by finding the objects}, and hence has an advantage for benchmarks where objects appear at random locations and over cluttered backgrounds. Hence, as expected, StarNet attains large performance gains (of $4\%$ and $5\%$ above SOTA baselines in $1$-shot setting) on CUB and ImageNetLOC-FS few-shot benchmarks, where images are less cropped around the objects. Notably, on these benchmarks we observe these gains also above the SOTA attention based and dense-matching based methods.
The results of the evaluation, together with comparison to previous methods, are given in Table \ref{table:results_classification}. Additional few-shot classification experiments showing StarNet's comparable performance on (cropped) \textit{mini}ImageNet and CIFAR-FS few-shot benchmarks are provided in Appendix.

\subsection{Ablation study} \label{sec:few-shot-classification-ablation}
%
We perform an ablation study to verify the contribution of the different components of StarNet and some of the design choices. We ablate using the $1$-shot, $5$-way CUB few-shot classification experiment, results are summarized in Table \ref{table:ablation}. 
To test the contribution of object detection performed by the StarNet (stage-1), we use the same global average pooling for the prototype features as in StarNet stage-2, only without weighting by $BP_{Q|S}$ and $BP_{S|Q}$ (\textit{'unattended stage-2'} in the table). We separately evaluate the performance of StarNet stage-1 and StarNet stage-2, this time stage-2 does use weighted pooling with $BP_{Q|S}$ and $BP_{S|Q}$. We then evaluate the full StarNet method (\textit{'full StarNet'}). As expected we get a performance boost as this combines the structured (geometric) evidence from stage-1 with unstructured evidence pooled from the object regions in stage-2. Finally, using the NMS process to iteratively extend the back-projected query region matched to the support attains the best performance.
\begin{table}[htb]
\begin{center}
\caption{\textbf{Ablation study} on CUB $1$-shot / $5$-way }\label{table:ablation}
\begin{tabular}{|l|c|}
\hline
unattended stage-2                           &  72.92  \\
StarNet stage-1	                               &  75.86  \\
StarNet stage-2	                           &  76.74  \\
full StarNet                       & 78.78  \\
full StarNet with iterative NMS  $\quad$    &  \textbf{79.58}  \\
\hline
\end{tabular}
\end{center}
\vspace{-1.5em}
\end{table}

\section{Conclusions}\label{sec:conclusions}
We have proposed a new Weakly-Supervised Few-Shot Object Detection (WS-FSOD) few-shot task, intended to significantly expedite building few-shot detectors for new visual domains, alleviating the need to obtain expensive bounding box annotations for a large number of base classes images in the new domain. We have introduced StarNet, a first WS-FSOD method. StarNet can also be used for few-shot classification, being especially beneficial for less-cropped objects in cluttered scenes and providing plausible explanations for its predictions by highlighting image regions corresponding to objects shared between the query and the matched support images. We hope that our work would inspire lots of future research on the important and challenging WS-FSOD task, further advancing its performance.
\bibliography{references}
\section{Appendix}
\begin{figure*}[ht!]
\begin{center}
   \includegraphics[width=1.0\textwidth,clip]{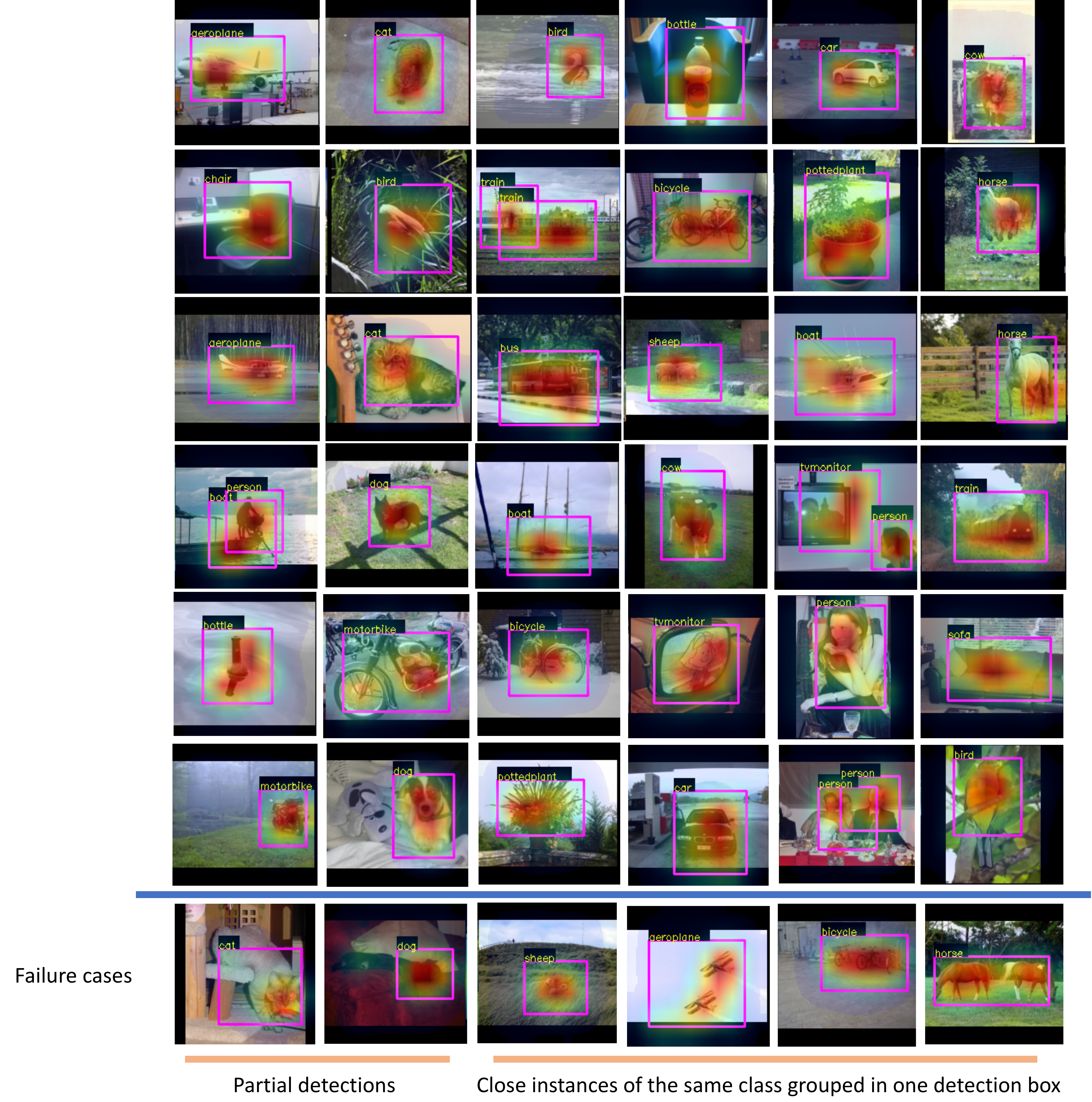}
\end{center}
   \caption{Example detections (on query images) from the PASCAL VOC WS-FSOD experiment described in the paper. Both detected object bounding boxes and a union of all their detected heatmaps produced by StarNet are visualized. Best viewed in color and in zoom.}
\label{fig:collage_pascal}
\end{figure*}
\subsection{Example PASCAL detections and example regions matched by StarNet}
\begin{figure*}[ht]
\begin{center}
   \includegraphics[width=1.11\textwidth,trim={0 0 9.5cm 0},clip]{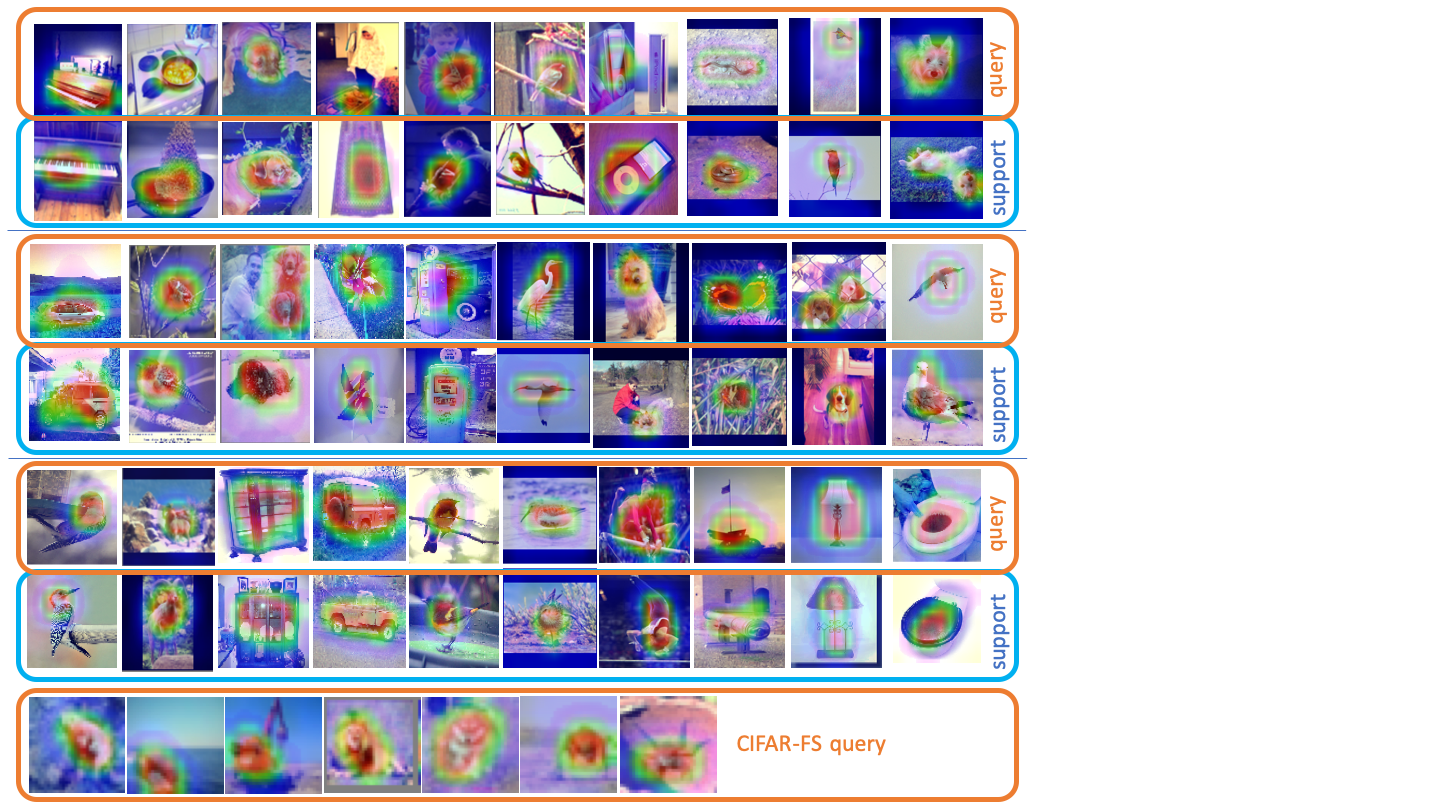}
\end{center}
   \caption{Examples of regions matched by StarNet between support and query images in few-shot episodes of different datasets. Top $3$ rows shows a mix of examples from \textit{mini}ImageNet, ImageNetLOC-FS, and CUB. Bottom row shows some query examples from CIFAR-FS with localized objects, despite the tiny $32 \times 32$ resolution. Best viewed in color and in zoom.}
\label{fig:collage}
\end{figure*}

Figure \ref{fig:collage_pascal} shows some detection examples and some failure cases from different episodes and different novel category sets used in the PASCAL VOC WS-FSOD experiments described in the paper. Some additional examples of regions matched during StarNet few-shot inference on different datasets are shown in figure \ref{fig:collage}.

\subsection{Videos depicting the evolution of the back-projection heatmaps during StarNet training}
In the video files attached to this supplementary we would like to demonstrate the evolution of StarNet`s localization capability, expressed via the back-projection heatmaps, evolving during the training process. In order to produce the visualization, we follow fixed 1-shot 5-way episodes, containing 4 query images per class, along the training epochs of the network. The images of these episodes come from the training categories, but are not used in the training.

Each video frame displays the 20 fixed query images of the selected episode (5 classes x 4 queries per class) on the left side, so that each row has 4 query images from the same class, and the 20 matching images from the support set on the right. The support set for 1-shot problem contains just 1 image per class, therefore the support images are expected to appear a number of times. Specifically, the support image presented at each location of the 4x5 grid is the one that matches best (according to the current state of the training network) to the query image in the corresponding grid location on the left. On both the query and support images we overlay the back-projection heatmaps generated for the query-support pairs as detailed in the paper.

One can observe in the videos that in the beginning of the training the masks are random and are not focused on any specific location, and the classifications are inaccurate (which is expressed by poor query-support matching). As the training progresses, the masks become smaller and more focused on the object, and the classification gets closer to being correct. Please note that as explained in the paper, only image level class labels supervision (without specifying object locations) is used for training. In the case of perfect classification, one should expect each row on the right side to be filled with the same support image (the only one available) of the correct class. 

The heatmaps evolution shows, therefore, how the StarNet's object localization capability evolves naturally with its classification ability.

The visualization videos were produces for several sample episode from two datasets: ImageNet-LOC and CUB. The contents for each episodes were not cherry-picked for easy matching, and the behavior observed in the videos is typical for the training process and across the datasets used in our experiments.
\begin{table*}[!ht]
	\centering
	\caption{Average precision (AP, \%) of weakly supervised few-shot detection and comparison to baselines on the ImagenetLOC-FS and CUB datasets. GC = GradCAM, SS = SelectiveSearch.}
	    \label{Table:results_det_IoP}
	\scriptsize{
	\begin{tabular}{llccc|ccc}
     \multicolumn{6}{c}{1-shot}  & \multicolumn{1}{c}{5-shot}  \\
    dataset & method   & \tiny{$IoU\geq0.3$} & \tiny{$IoU\geq0.5$}  & \tiny{$IoP\geq\dfrac{2}{3}$} &  \tiny{$IoU\geq0.3$} & \tiny{$IoU\geq0.5$} &
    \tiny{$IoP\geq\dfrac{2}{3}$} \\
    \shortstack{\textbf{Imagenet} \\ \textbf{-LOC}} &
    \shortstack{MetaOpt+GC} & 32.4       & 13.8      & 29.2         &  51.9   &   22.1   &    41.4   \\
     & \shortstack{MetaOpt+SS} & 16.1      & 4.9     & 6.7        &  27.4   &  10.2   &    12.7    \\
    & \shortstack{PCL \cite{PCL}} & 25.4      & 9.2    & 23.8    &  37.5  &   11.3   &    34.3  \\
    & \shortstack{CAN \cite{hou2019a}} & 23.2          & 10.3      & 20.1     &  38.2   &   12.7   &    35.1    \\
    & \textbf{StarNet (ours)}  & \textbf{50.0}       & \textbf{26.4}     & \textbf{43.6}        &  \textbf{63.6}    &   \textbf{34.9}  &    \textbf{54.8}    \\
    \textbf{CUB} &
        \shortstack{MetaOpt+GC} & 53.3          & 12.0      & 52.5        &  72.8   &   14.4   &    62.6    \\
        & \shortstack{MetaOpt+SS} & 19.4      & 6.0      & 7.8     &  26.2   &   6.4   &    4.2  \\
       & \shortstack{PCL \cite{PCL}} & 29.1      & 11.4    & 29.0    &  41.1  &   14.7   &    37.0    \\
        & \shortstack{CAN \cite{hou2019a}} & 60.7          & 19.3      & 55.4      &  74.8   &   26.0   &    66.1    \\
        & \textbf{StarNet (ours)}  & \textbf{77.1}        & \textbf{27.2}      & \textbf{71.4}       &  \textbf{86.1}   &   \textbf{32.7}   &    \textbf{78.7}    \\
	\end{tabular}
	}

\end{table*}
\subsection{Failure cases analysis - \textbf{partial detections} }
A common weakness of WSOD methods, shared also by our method, is that the predicted bounding boxes cover only a part of the object, usually the most salient one. In the scenario where one is interested in merely pointing at objects, rather than exactly bounding them, the $IoU \geq 0.5$ matching criteria (common in fully-supervised detection) is too restrictive. To analyze whether the performance drop in AP observed for all methods (StarNet and all the baselines) when moving from $IoU \geq 0.3$ to $IoU \geq 0.5$ results from partial detection, we consider the following pair of related measures. For a ground truth (GT) bounding box $G$ and a predicted box $P$ we define  $IoP=\dfrac{G\cap P}{P}$ (\textit{Intersection over Predicted}) computing the portion of the predicted box area covered by the GT box, and  $IoG=\dfrac{G\cap P}{G}$ (\textit{Intersection over Ground Truth}), computing the portion of GT box covered by the predicted box. Thus, IoP and IoG provide the precision and recall information, respectively, for object coverage. 

In the special case of equal-sized GT and predicted boxes, the $IoU=0.5$ corresponds to $IoP=\frac{2}{3}$. We use this intuition to substitute the $IoU \geq 0.5$ criterion with $IoP \geq \frac{2}{3}$ criterion, as a criterion better accounting for partial detection when computing the Average Precision (AP). The values of AP for $IoP \geq \frac{2}{3}$, for StarNet and the baselines, are provided in Table \ref{Table:results_det_IoP}. 

The AP of StarNet, using $IoP=\frac{2}{3}$, is substantially higher than that computed for $IoU \geq 0.5$, corroborating our suggestion that the performance drop between $IoU\geq0.3$ and $IoU\geq0.5$ is mostly due to partial detections. 
Additionally, we computed the average value of IoG \textit{only} for the boxes that passed the $IoP \geq \frac{2}{3}$ criterion and had the correctly predicted class label. This complements the picture, providing the average portions of GT objects covered by the (good) partial detections. We found that the StarNet bounding boxes that pass $IoP=\frac{2}{3}$ and have correct predicted class label still cover a significant portion of more than $32\%$ of the GT boxes for objects on average.

As can be seen from the table, the baseline methods attain considerably lower (than StarNet) AP values for $IoP \geq \frac{2}{3}$, consistent with the gains in performance observed for StarNet using the $IoU$-based criteria. This is likely due to StarNet being directly end-to-end optimized for classifying images by localizing the objects within them (using the proposed star-model geometric matching), while the other methods are either: not intended for few-shot (PCL), or optimized attention for classification and not for localization (CAN), or intended for classification and not localization (MetaOpt) - which cannot be easily bridged using the standard techniques for localization in classifiers (GradCAM, SelectiveSearch).

\subsection{Additional few-shot classification experiments}
While our paper is about Weakly-Supervised Few-Shot Object Detection (WS-FSOD), we have also conducted experiments to evaluate StarNet's few-shot classification performance. As described in the main paper, on the CUB and ImageNetLOC-FS benchmarks with \textit{non-cropped} objects appearing at random locations on cluttered backgrounds, StarNet demonstrates an advantage due to its ability to detect and focus on objects, even for unseen test classes. Intuitively, this allows better handling of fine-grained cases (CUB bird species) and background clutter (both). On two additional few-shot classification benchmarks: \textit{mini}ImageNet and CIFAR-FS, StarNet demonstrates essentially comparable (yet not superior) performance compared to other methods. We believe this is due to these datasets containing mostly cropped objects and hence StarNet's ability to detect objects plays a little role for them. The results including all the evaluated few-shot benchmarks are provided in table \ref{table:results_classification_appendix}.

\subsubsection{Additional datasets description}
The \textbf{\textit{mini}ImageNet benchmark} \cite{Vinyals2016} consists of $100$ classes from ILSVRC-2012 \cite{imagenet} split into $64$ meta-training, $16$ meta-validation, and $20$ meta-testing classes. Each class has 600 $84 \times 84$ images. The \textbf{CIFAR-FS} dataset \cite{Bertinetto2019} is built from CIFAR-100 \cite{Krizhevsky2009}. Its $100$ classes are split into $64$ training, $16$ validation and $20$ testing. Each class contains $600$ $32 \times 32$ images.

\begin{table*}[!ht]
  \centering
    \caption{Few-shot classification accuracy ($\%$), for all methods the $0.95$ confidence intervals are $<1\%$ (omitted for brevity). For fair comparison, showing only results that \textit{do not} use the validation set for training, \textit{do not} use the transductive or semi-supervised setting, use standard input resolution $84 \times 84$, 
    and \textit{do not use} additional information such as class label or class attributes embedding.
    \\$^{(1)}$Results from \cite{closer_look}, best result among resnet-10/18/34. 
    \\$^{(2)}$using official code and best hyper-parameters between defaults and those found by tuning on validation set for each benchmark.
    \\$^{(3)}$ shared backbone between StarNet stage-1 and stage-2.
    \\$^{(4)}$ also used validation data for training.
    \\$^{(5)}$DeepEMD trained and tested on \textit{cropped} CUB, where all birds are cropped using ground truth bounding boxes before being down-sized to $84\times84$. In contrast, StarNet is trained and tested on \textit{non-cropped} CUB with birds appearing at random locations on cluttered backgrounds, with entire images downsized to $84\times84$. 
    \\$^{(6)}$according to the publicly released code of DeepEMD, its train data loader uses random crops from high (more than $224\times224$) resolution images resized to $84\times84$ (StarNet and other methods only have access to full images at $84\times84$ resolution), and its test data-loader uses $10\%$ higher image resolution than StarNet and other methods.
  }\label{table:results_classification_appendix}
  \scriptsize{
  \begin{tabular}{lccccccccc}
      & &
      \multicolumn{2}{c}{ImageNetLOC-FS} &
      \multicolumn{2}{c}{CUB} & 
      \multicolumn{2}{c}{\textit{mini}ImageNet} & 
      \multicolumn{2}{c}{CIFAR-FS} \\
      \textbf{method} & 
      backbone &
      1-shot & 5-shot & 
      1-shot & 5-shot & 
      1-shot & 5-shot & 
      1-shot & 5-shot 
      \\
   Baseline$^{(1)}$ \cite{closer_look}
   & resnet-34$^{(1)}$
   & - & - & 67.96 & 84.27 & 52.37 & 74.69 &-&-
   \\
    Baseline++$^{(1)}$ \cite{closer_look}
   & resnet-34$^{(1)}$
     & - & - & 69.55 & 85.17 & 53.97 & 76.16 &-&-
    \\
    MatchingNet$^{(1)}$ \cite{Vinyals2016} 
   & resnet-34$^{(1)}$
     & - & - & 72.36 & 83.78 & 54.49 & 68.88 &-&-
    \\
    MAML$^{(1)}$ \cite{Finn2017}
   & resnet-34$^{(1)}$
     & - & - & 72.36 & 83.78 & 54.69 & 66.62 &-&-
    \\
   ProtoNet$^{(1)}$ \cite{Snell2017}
   & resnet-34$^{(1)}$
     & - & - & 72.03 & 87.42 & 54.16 & 74.65 &-&-
    \\
    RelationNet$^{(1)}$ \cite{relationnet} 
   & resnet-34$^{(1)}$
    & - & - & 68.65 & 82.75 & 52.48 & 70.2 &-&-
    \\
   SAML \cite{Hao2019} 
                & conv4
                  & - & - 
                  & 69.35 & 81.56
                  &  57.69 &  73.03
                  & -  & -
                  \\
   FSL with Loc. \cite{Wertheimer2019} 
                  & conv4
                  & - & - 
                  & - &  -
                  & 51.1 & 69.45
                  & - & -
                  \\
   TADAM \cite{Oreshkin2018}
                  & resnet-12
                  & - & - 
                  & - & - 
                  & 58.50 & 76.70
                  & -  & -
                  \\
  LEO$^{(4)}$ \cite{leo} 
                 & wrn-28-10
                  & - & - 
                  & - & - 
                  & 61.76 & 77.59
                  & -  & -
                  \\
   Variat. FSL \cite{Zhang2019}
                  & resnet-12
                  & - & - 
                  & - &  -
                  & 61.23 & 77.69
                  & - & -
                  \\
  wDAE-GNN \cite{Gidaris2019}
                  & wrn-28-10 
                  & - & - 
                  & - & - 
                  & 61.07 & 76.75
                  & -  & -
                  \\
   CTM \cite{Li2019}
                  & resnet-18
                  & - & - 
                  & - & -
                  &  64.12 &  80.51
                  & -  & -
                  \\
  CC+rot \cite{Gidaris2019}
                  & wrn-28-10 
                  & - & - 
                  & - & - 
                  & 62.93 & 79.87
                  & 73.62  & \textbf{86.05}
                  \\
   Dist. ensemble \cite{Dvornik2019}
                  & ensemble of $20\times$ resnet18
                  & - & - 
                  & 68.77 & 83.57 
                  & 59.38 & 76.9
                  & -  & -
                  \\       
   $\Delta$-encoder \cite{Schwartz2018}
                & resnet-18
                & - & - 
                & 69.80 & 82.60
                & 59.90 & 69.70
                & 66.70 & 79.80
                \\       
  MTL \cite{sun2019a}
                & resnet-12
                & - & - 
                & - & -
                & 61.2 & 75.5
                & -  & - 
                \\   
  DC \cite{lifchitz2019a}
                & resnet-12
                & - & - 
                & - & -
                & 61.26 & 79.01
                & -  & -
                \\   
   CAN \cite{hou2019a}
                & resnet-12
                & 57.1\tiny{$^{(2)}$} & 73.9\tiny{$^{(2)}$} 
                & 75.01\tiny{$^{(2)}$} & 86.8\tiny{$^{(2)}$}
                & 63.85 & 79.44
                & - & -
                \\       
   MetaOpt \cite{Lee2019}
                  & resnet-12
                  & 57.7\tiny{$^{(2)}$} & 74.8\tiny{$^{(2)}$} 
                  & 72.75\tiny{$^{(2)}$} & 85.83\tiny{$^{(2)}$}
                  & 62.64 & 78.63
                  & 72.6  & 85.3
                  \\
DeepEMD$^{(6)}$ \cite{Zhang2020a}
                  & resnet-12
                  & - & -
                  & 75.65$^{\textbf{(5)}}$ & 88.69$^{\textbf{(5)}}$ 
                  & \textbf{65.91$^{\textbf{(6)}}$} & \textbf{82.41$^{\textbf{(6)}}$} 
                  & - & - \\
\hline
   \textbf{StarNet - shared backbone (ours)}\tiny{$^{(3)}$}
                  & resnet-12
                  & 61.0 & 77.0 
                  & 79.44 &  88.8 
                  & 61.8 & 78.5
                  & \textbf{73.65} & 85.9 \\
   \textbf{StarNet (ours)}
                  & $2\times$ resnet-12 = resnet-18 
                  & \textbf{63.0} & \textbf{78.0} 
                  & \textbf{79.58} &  \textbf{89.5} 
                  & 63.4 &  80.3
                  & 73.34 & 85.95
                  \\        
\end{tabular}
}
  \vspace{-2em}
\end{table*}

\end{document}